\documentclass[10pt, a4paper]{article}

\usepackage{graphicx}
\usepackage{caption}
\usepackage{subcaption}
\usepackage{booktabs}
\usepackage{multirow}
\usepackage{amsmath}
\usepackage{colortbl} 
\usepackage{rotating}

\usepackage{hyphenat}
\definecolor{lightgray}{gray}{0.8}

\usepackage{lrec-coling2024} 

\title{The Role of Syntactic Span Preferences in Post-Hoc Explanation Disagreement}

\name{Jonathan Kamp$^1$, Lisa Beinborn$^{1,2}$, Antske Fokkens$^{1,3}$} 

\address{$^1$Computational Linguistics and Text Mining Lab, Vrije Universiteit Amsterdam \\
$^2$Institute of Computer Science and Campus Institute Data Science, University of Göttingen \\
$^3$Dept. of Mathematics and Computer Science, Eindhoven University of Technology \\
         \{j.b.kamp, antske.fokkens\}@vu.nl, lisa.beinborn@uni-goettingen.de\\}
         
\abstract{
Post-hoc explanation methods are an important tool for increasing model transparency for users. Unfortunately, the currently used methods for attributing token importance often yield diverging patterns. In this work, we study potential sources of disagreement across methods from a linguistic perspective. 
We find that different methods systematically select different classes of words and that methods that agree most with other methods and with humans display similar linguistic preferences. Token-level differences between methods are smoothed out if we compare them on the syntactic span level. We also find higher agreement across methods by estimating the most important spans dynamically instead of relying on a fixed subset of size $k$. 
We systematically investigate the interaction between $k$ and spans and propose an improved configuration for selecting important tokens.
 \\ \newline  }

\begin{document}

\maketitleabstract

\section{Introduction}
\begin{figure}[htbp]
  \centering
    \includegraphics[width=0.48\textwidth]{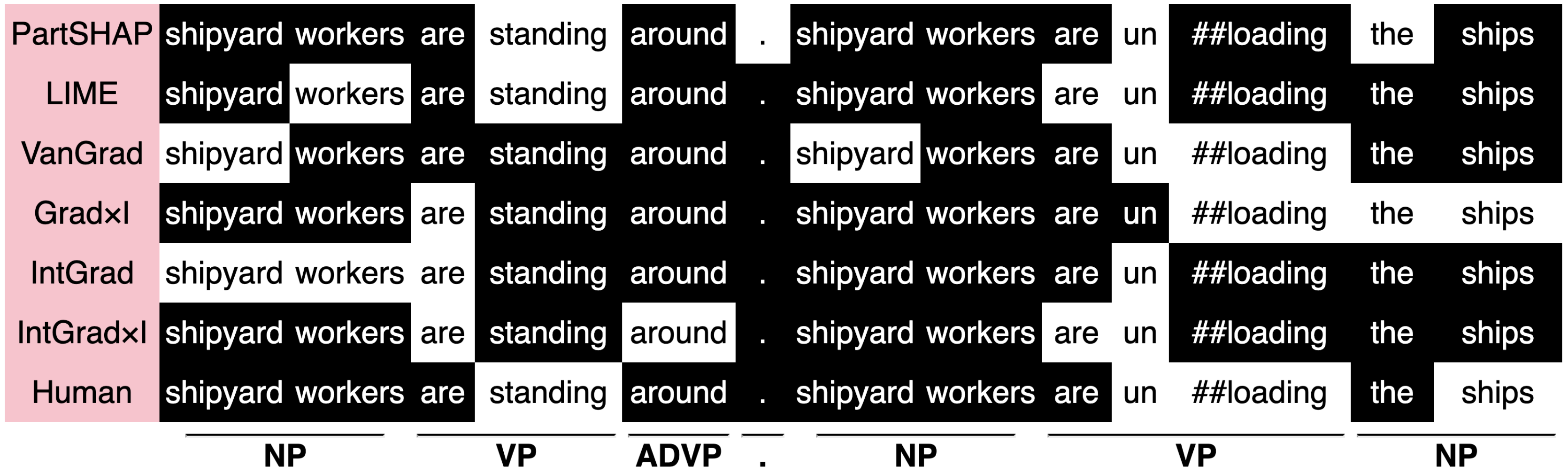}
  \caption{Top-$k$ highlights (light background) per attribution method and human preference for $k=4$. The syntactic spans are given underneath.}
  \label{fig:figure1}
\end{figure}

Transformer-based models learn to map features in the input to some output. 
When training an NLP system, the model learns to identify the most important features (in our case tokens) for the final prediction. Post-hoc explanation methods such as LIME \citep{ribeiro-etal-2016-trust} and Integrated Gradient \citep{sundararajan2017axiomatic} aim to attribute an importance score to the individual features to interpret the model's decisions. Generally, these methods tend to disagree with each other when ranking token importance on a set of top-$k$ tokens based on attribution scores \citep{neely2022song}. Given their disagreement, and assuming that explanations that are faithful to the transformer’s inner mechanisms should be agreeable \citep{jain2019attention}, the faithfulness of these methods comes under question. However, methods might agree more than initially appears. 
For example, Figure \ref{fig:figure1} shows that none of the methods selects the same top-4 tokens and that 12 of the 13 tokens appear in at least one top-4 selection, indicating a high variance across 
methods. 
Intuitively though, methods seem to target the verb phrases \textit{are standing} and \textit{are unloading} to a high degree as the vast majority highlights at least one of the tokens in each of these phrases.
Similarly, some methods tend to agree on the noun phrases \textit{shipyard workers} (first occurrence) and \textit{the ships}, and even more so on different tokenised subwords of the same word, namely \textit{un} and \textit{\#\#loading}. 
This leads us to hypothesise that agreement between methods is systematically higher when we look at the linguistic spans they are targeting: the constituents to which tokens syntactically belong. 

This example shows that a single method may have a specific \textit{preference} for one word class over another, e.g.\ noun over adjective, auxiliary over inflected verb form or modifier over head. For example, \citet{ramnath2020towards} report part-of-speech (POS) preference statistics for the different layers of BERT \citep{devlin-etal-2019-bert} for the Integrated Gradient method. However, the extent to which preferences differ across methods remains unclear, as well as its impact on method--method agreement.

A methodological aspect that directly affects agreement is the selection of the top-$k$ most important tokens for each method to compare. $k$ is a relatively under-explored parameter and is defined as the number of features that are assigned highest scores by the attribution method, relative to all the features in the input example. A common way of picking $k$ is by selecting a fixed number, generally in the range $[1, 10]$. Intuitively, a $k$ that is fixed across instances (e.g.\ 4) is suboptimal, and the selection process of $k$ is often overlooked \citep{jesus2021can, camburu2019can} or obtained by an approximation \citep{krishna2022disagreement}. As an alternative, $k$ can be estimated dynamically across instances \citep{pruthi2022evaluating, kamp-etal-2023-dynamic}, but different conceptual settings for this approach and their effect on agreement have not been investigated yet.
Instead of ranking tokens by attribution score and manually setting a $k$, \citet{kamp-etal-2023-dynamic} propose to automatically detect tokens that are signal peaks in the input. Hypothesising that spans are better suited for agreement than tokens conceptually overlaps with this dynamic $k$ approach. Precisely, the latter suggests that solely focusing on token-level attribution scores, semi-arbitrary importance cut-offs and the consequent agreement measurements between $k$ tokens may be undesirable for interpreting model behaviour.

In this paper, we aim to disentangle the interdependencies between word class preference, span-level agreement, and the determination of $k$. We show that methods systematically select different word classes and that methods that agree most with other methods and with humans exhibit similar word class preferences. We also find that dynamic $k$ and spans work well in combination, and that an adapted threshold for dynamically selecting the $k$ most important tokens passes our baseline tests for both token- and span-level $k$ estimation. Our main contributions are: i) a linguistic analysis of disagreement on the token-level and on the span-level and ii) an improvement to the dynamic-$k$ estimation algorithm.\footnote{All analyses are available at: \\ \url{https://github.com/jbkamp/repo-Span-Pref}}

\section{Related Work}
In this section, we place our work in the context of prior work on interpretability (§\ref{sec:model_interpretation}), the patterns of linguistic information that attribution methods reveal (§\ref{sec:encoding_linguistic_information}) and top-$k$ estimation (§\ref{sec:topk_estimation}).

\subsection{Model Interpretation}\label{sec:model_interpretation}
Tracing the decision processes in neural models poses difficulties due to various factors, including their non-linear nature and the absence of explicit human-defined rules to link patterns in the input features with output labels. Different research lines exist to interpret different aspects of the model \citep{choudhary2022interpretation, rauker2023toward}, such as the linguistic information that might implicitly be learned by the model, or the importance that single input features might have had towards the model's decision \citep{madsen2022post}. 

To address the latter, post-hoc attribution methods in NLP have been developed to assign a score to each token in the input, creating an \textit{attribution profile} over the tokens. While these methods are often being used in error analyses \citep[i.a.]{bongard2022legal}, their reliability is questionable. In fact, attribution profiles obtained from different methods can differ strongly even on the same input. This leads to an overall low inter-method agreement \citep{neely2022song}, which has also been found for domains outside of NLP \citep{krishna2022disagreement}. Diverging experimental results of such methods on different models, datasets and tasks provide additional evidence on their inconsistency. For example, when trying to identify the attribution methods that best align with human preferences--the most \textit{plausible} \citep{jacovi2020towards} methods--, \citet{atanasova2020diagnostic} and \citet{attanasio2022benchmarking} come to fundamentally opposing conclusions. \citet{roy2022don} characterise disagreement between methods in a software defect prediction task as being highest in terms of top-$k$ feature importance, followed by rank, then sign. Similarly to \citet{pirie2023agree}, they propose aggregation schemes for different explanation methods that aim to tackle disagreement in real-world use cases. 

One question that, to our knowledge, remains under-explored, is \textit{why} attribution methods in NLP disagree. A key to answering this would be comparing methods on their \textit{faithfulness}, i.e.\ the degree to which methods are reflecting the model's decision making process, as recent work \citep[i.a.]{atanasova-etal-2023-faithfulness} aims to assess. However, directly measuring faithfulness might only find glimpses of the model's inner workings rather than providing a conclusive answer \citep{jacovi2020towards}. Therefore, we think that the first step should be explaining disagreement by the observable output of the methods, i.e.\ the attribution profiles. We aim to provide a linguistic comparison by quantifying the kind of features that are targeted, expecting different methods to consistently target different classes of words.

\subsection{Linguistic Patterns in Attributions}\label{sec:encoding_linguistic_information}
Identifying the linguistic preferences of models is important in order to pinpoint the cues upon which models depend during inference time. Only a handful of studies have explored POS preference. Especially in a feature attribution setting, there is little evidence that shows certain preferences by different attribution methods and how these preferences differ. \citet{lai2019many} find that different \textit{models} (i.e.\ LSTM, XGBoost and SVM) have different POS preferences on the same data and task, but they do not explore preferences for different \textit{attribution methods}. \citet{ramnath2020towards} examine the top-5 most important tokens in each layer and find that BERT \citep{devlin-etal-2019-bert} primarily focuses on nouns in all 12 layers, followed by verbs and adjectives. Interestingly, both punctuation tokens and stop words each correspond to 10\% in the top-5 selections. However, only Integrated Gradient \citep{sundararajan2017axiomatic} was used in this experiment, limiting the generalisability of their findings. Our analyses differ from theirs in that we compare different methods and investigate the overlap between agreement and linguistic preference. 

Language (and model behavior) can often not be explained by merely highlighting individual tokens. Rather, we would ideally observe how features act in combination with each other and, for example, if they do so hierarchically. As an alternative way of analysing the attributions of tokens in isolation, we find a growing line of research on feature \textit{interactions}. \citet{jumelet-zuidema-2023-feature} find evidence of attribution methods faithfully reflecting linguistic structure in language models. \citet{sikdar2021integrated} combine token-wise attribution scores into scores assigned to syntactic parent constituents. Similarly, \citet{babiker2023intermediate} train a model on intermediate representations in a hierarchical fashion. \citet{song2023automatic} aim to capture the causal effect of word group combinations on the prediction but limit their scope to the Integrated Gradient method.  \citet{pruthi2022evaluating} anticipate that certain spans of tokens should be highlighted by attribution methods in a sentiment analysis task.
While their intuition is on point, the relatively broad expectations found in the latter underscore the relevance of a clear definition of token spans and their role in demonstrating how neighboring features are grouped.

As far as we know, there is no prior work that covers a linguistic analysis of the token selections targeted by different attribution methods. To the best of our knowledge, we are also the first to investigate the relation between disagreement on the linguistic level to overall disagreement among methods. We provide a linguistic analysis in terms of individual tokens, and also in terms of spans that have a clear syntactic definition. In particular, we link disagreement to linguistic preference on the token level and within spans.

\subsection{Top-$k$ Estimation}\label{sec:topk_estimation}

We analyse the factors of disagreement through an additional scope, namely top-$k$ estimation. $k$ represents the number of most important tokens in the attribution profile. Studies reporting on consistent disagreement between methods do not take the impact of the $k$ number of selected tokens into account \citep{pruthi2022evaluating, krishna2022disagreement, neely2022song}. A common way of selecting $k$ is approximating it to a \textit{low} value, e.g.\ 1 or 2 \citep{bastings-etal-2022-will}, 5 \citep{ramnath2020towards}, 5 or 10 \citep{camburu2019can}, 25\% of the average input length \citep{krishna2022disagreement}. 
However, a $k$ that is fixed does not account for variability among instances. A $k$ that is too low can exclude important tokens from the comparison, whereas a $k$ that is too high will include non-important tokens while artificially boosting agreement between methods. Keeping $k$ relatively low also helps users to more easily digest the explanations in a real-world scenario.

The value of $k$ has also been estimated dynamically. \citet{pruthi2022evaluating} set $k$ to 10\% of the input length, assuming that longer inputs have a higher number of important features than shorter inputs. \citet{kamp-etal-2023-dynamic} propose a $k$ that varies dynamically based on properties of the attribution profile of each instance, aiming to include features that display above average importance and that focus more on the targeted region of the input instead of the specific token. While their method estimates a value for $k$ that is close to human preference, we find that their algorithm necessitates further experiments and refinement. Different importance thresholds are possible and need baseline benchmarking. Also, as of now, prior methods for determining dynamic $k$ do not explicitly account for negative attribution scores.

We adopt and improve the \textbf{dynamic $k$} estimation by \citet{kamp-etal-2023-dynamic} throughout §\ref{sec:agreement_at_the_span_level}, when measuring agreement at the span level compared to the token level. Formally, this dynamic approach defines a strong signal in the attribution profile as a score that is higher than its neighboring scores according to two principles: \textit{local importance} and \textit{global importance}. \textit{Local importance} requires that a score must be higher than its strict neighbors ($\pm1$ window) to reduce redundancy of tokens belonging to the same signal. In other words, a set of adjacent tokens with relatively high scores is converted to a single important signal and the highest attribution in the set is kept as the peak of the signal. Similarly, the \textit{global importance} principle requires important signals to be minimally above average signal strength, i.e.\ $>\mu_{ap}$, where $ap$ is the attribution profile. By only adopting the \textit{global importance} threshold, the inclusion of groups of (redundant) neighboring tokens with high attribution scores is expected to increase $k$, unnecessarily boosting the agreement scores. Therefore, the addition of a \textit{local importance} setting, which we keep unaltered for our remaining experiments, is necessary to estimate signal \textit{peaks}. As for \textit{global importance}, we keep the threshold constant in §\ref{sec:effectdynkonspans} to compare span-level agreement to token-level agreement in previous work, and explore different settings in §\ref{sec:differentdynksettings}. 


\section{Linguistic Analysis}\label{sec:Linguistic_analysis} 
We hypothesize that one of the reasons attribution methods disagree is that different methods have different preferences for the classes of words they target. Following from this, we expect that differences in word class preferences are put under a different light when we look at the syntactic spans they are assigned to.

\subsection{Setup}
To analyse the disagreement problem, we consider six different attribution methods on a natural language inference task. For the sake of testing our hypothesis against the agreement results from prior work, we follow \citet{kamp-etal-2023-dynamic} for the experimental setup. For the backbone model, we use the default training split (549,361 instances) of the e-SNLI dataset \citeplanguageresource{camburu2018snli} to finetune DistilBERT \citep{sanh2019distilbert} 10 times on 10 different random seeds. We then use the model (0.89 F1) with the least variation in attribution profiles on the default test split (9,842 instances) for analysis. One instance in the dataset corresponds to the concatenation of a premise followed by a hypothesis. The possible output labels are contradiction, entailment and neutral, making it a multi-class problem. Classes are balanced and indicate the relation between premise and hypothesis. 

The words in every instance are also annotated as being important or not important towards the output label (3~annotators per instance, 4$\pm$3 important words on average), producing so-called human \textit{rationales} \citep{carton-etal-2020-evaluating}. From these human rationales, we derive word-level aggregation scores comprised in the interval $[0, 1]$ indicating the proportion of annotators that found the word important. These scores are used to compare attribution scores to human preference when considering a top-$k$ selection (see \textit{Human} in Figures~\ref{fig:figure1}, \ref{fig:freq_keq4} and~\ref{fig:pairwise_chunk_fixedvsdyn}). As for the attribution methods, we use both \textit{gradient-based} approaches by including Vanilla Gradient \citep{simonyan2014deep}, Integrated Gradient \citep{sundararajan2017axiomatic}, and both versions multiplied with the input \cite{shrikumar2017learning}, as well as \textit{perturbation-based} approaches, by including Partition SHAP \citep{lundberg2017unified} and LIME \citep{ribeiro-etal-2016-trust}.\footnote{Ferret package v0.4.1 \citep{attanasio-etal-2023-ferret}.}

\subsection{Preference for a Word Class}\label{sec:token_type_preference}
\begin{figure}
\centering
\begin{subfigure}[b]{\linewidth}
  \centering
  \includegraphics[width=1.0\linewidth]{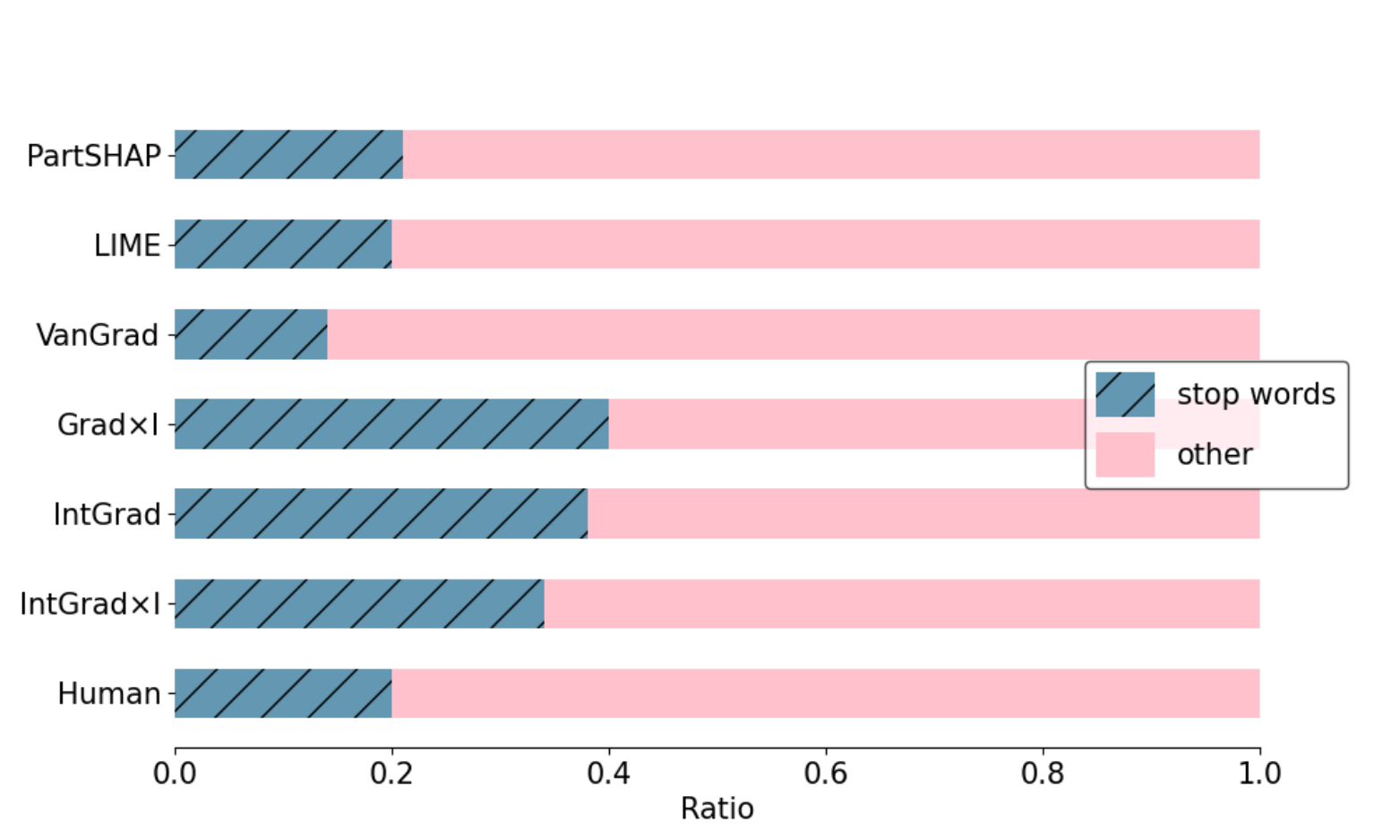}
  \caption{Relative frequency for important stop words, $k=4$.}
  \label{fig:freq_stop_keq4}
\end{subfigure}
\begin{subfigure}[b]{\linewidth}
  \centering
  \includegraphics[width=1.0\linewidth]{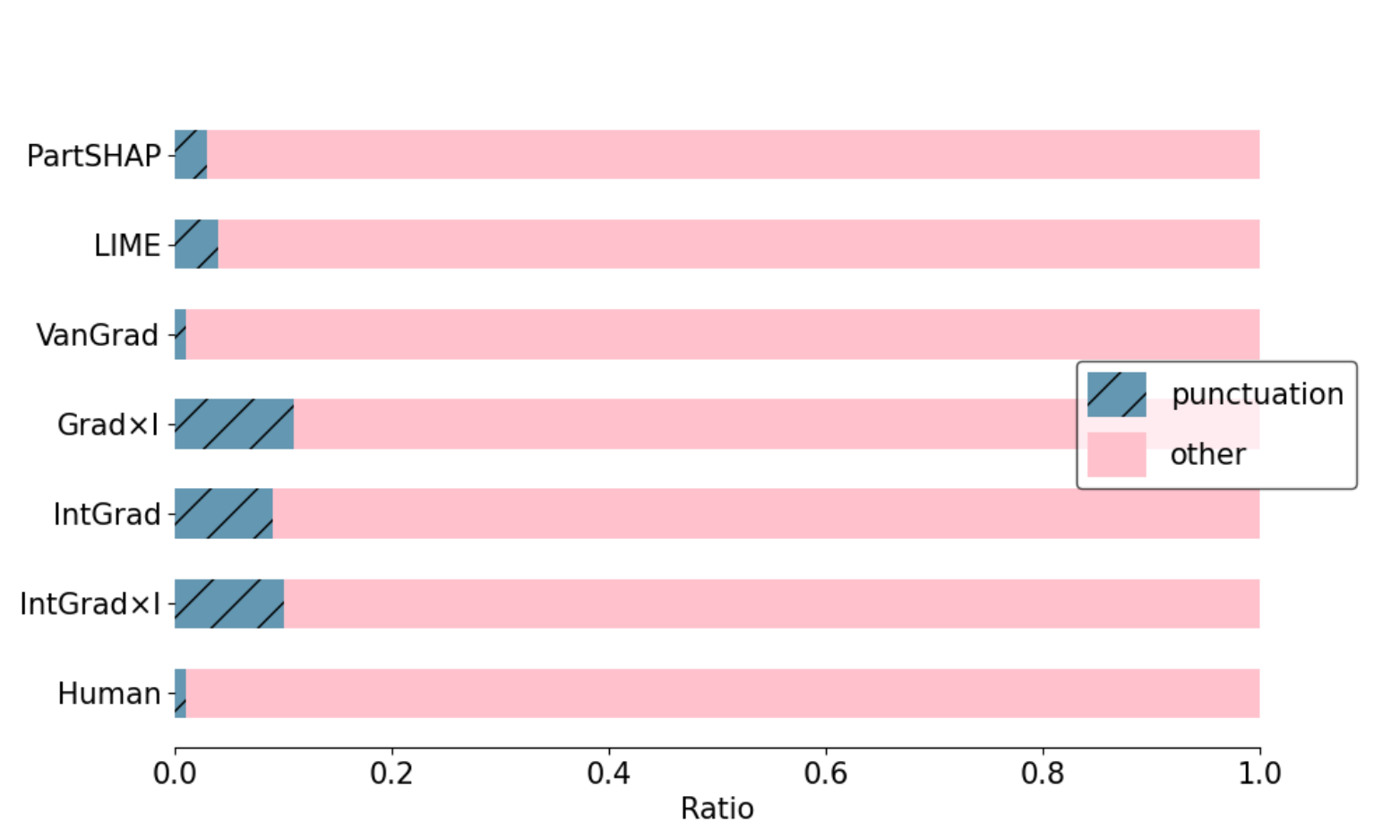}
  \caption{Relative frequency for important punctuation, $k=4$.}
  \label{fig:freq_punct_keq4}
\end{subfigure}
\begin{subfigure}[b]{\linewidth}
  \centering
  \includegraphics[width=1.0\linewidth]{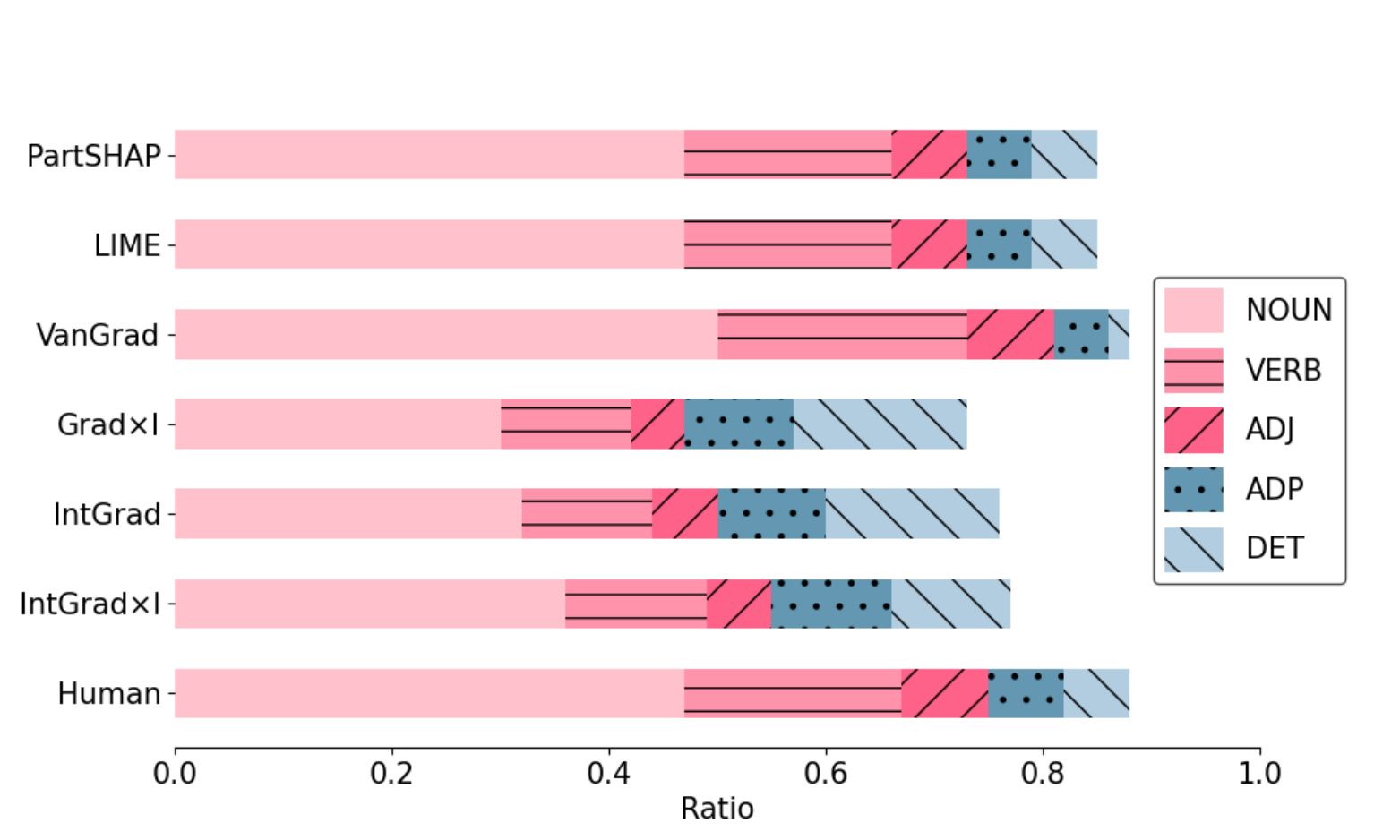}
  \caption{Relative frequency for important POS, $k=4$. We consider the 5 most preferred POS tags by humans.}
  \label{fig:freq_pos_keq4}
\end{subfigure}
\caption{Preference for different word classes per attribution method.}
\label{fig:freq_keq4}
\end{figure}

The first step in our analysis compares word class preference of different attribution methods on top-$k$ tokens.
We set $k$ to 4 which corresponds to the average number of tokens that were highlighted by humans in e-SNLI. This value is reflected by a comparable value of averaged dynamic $k$ and comparable method--method agreement levels as found by \citet{kamp-etal-2023-dynamic}. Figure~\ref{fig:freq_keq4} illustrates the occurrence of different word classes among the tokens with the highest attribution values (i.e.\ \textit{important} tokens) for each method and for human aggregated annotations.
We compare the ratio of important stop words (Figure \ref{fig:freq_stop_keq4}), punctuation tokens (\ref{fig:freq_punct_keq4}), and the distribution of the five most preferred POS tags by humans: \textsc{noun}, \textsc{verb}, \textsc{adj}, \textsc{adp}, \textsc{det} (Figure \ref{fig:freq_pos_keq4}).
Interestingly, with regards to Integrated Gradient, Gradient\,×\,Input and Integrated Gradient\,×\,Input, roughly 10\% in each top-4 selection on average consists of punctuation. Despite question answering and natural language inference being different tasks, we replicate the findings on punctuation preference for Integrated Gradient by \citet{ramnath2020towards}. Notably, these findings do not generalise to the other methods.

Intuitively, this preference seems to be inherent to the method and not to the underlying model, as each instance normally is a concatenation of two sentences tailed by a full stop each; hence, it is very unlikely that the model is using punctuation as shortcut signals to the output labels. This might suggest that some methods pick up information about the approximate location of a signal in the sentence (\textit{locality} information), rather than the precise token (\textit{lexical} information).
While punctuation may be a simple symptom of locality, it is important to further examine this phenomenon in the broader context of spans. We do so through a linguistic analysis of spans of locally adjacent tokens, the use of dynamic $k$, and their intersection in §\ref{sec:agreement_at_the_span_level}.

Stop words on the other hand do not display a similar preference as found by \citet{ramnath2020towards} (40\% versus 10\% for Integrated Gradient), indicating that this difference might be task-related. For the other POS tag preferences, we do not observe a clear overlap with prior research for Integrated Gradient (\textsc{noun}: no overlap; \textsc{verb}: overlap; \textsc{adj}: no overlap; \textsc{adp}: cannot compare; \textsc{det}: cannot compare). What we do observe from Figure~\ref{fig:freq_keq4}, is the systematic different preference for stop words, punctuation and most frequent POS tags by Integrated Gradient, Gradient\,×\,Input and Integrated Gradient\,×\,Input (Group 1), compared to the other methods and to humans (Group 2). Hence, this intuitively leaves us with two groups displaying different word class preferences.

Assuming that methods (including human rationales) are independent, we apply Chi-Square tests to method--method (and human--method) pairs' preference distributions.\footnote{The full Chi-Square tests are given in Appendix \ref{appendix}.} For each pair, we measure whether there is a significant difference between stop word distributions, between punctuation distributions and between POS tag distributions. The tests confirm our initial observations that most distributions from one group are significantly different from the other group (25/36 pairs,\footnote{A total of twelve Group 1 -- Group 2 comparisons are possible for each of the three word classes (stop words, punctuation and POS), resulting in 36 pairs.} with $p<.05$) and that no significant differences are found within groups. Most of the exceptions arise for pairs involving Integrated Gradient\,×\,Input, with 3 out of 3 non-significant differences found in combination with Partition SHAP,\footnote{For each pair of methods, there are three word classes for which significance can be tested. We therefore compute the ratio '$i$ out of 3'.} 2 out of 3 with LIME and 1 out of 3 with human rationales. Hence, Integrated Gradient\,×\,Input explains half (6/11) of the non-significant differences found and can roughly be placed in between the two groups. Additionally, punctuation preferences account for half (6/11) of the non-significant differences between groups. This might be due to the small numbers of the punctuation frequencies, which may have affected the Chi-Square statistics.

Primarily Integrated Gradient and Gradient\,×\,Input, followed by Integrated Gradient\,×\,Input, are indeed the methods for which \citet{kamp-etal-2023-dynamic} find that method--method and human--method agreement are lowest. 
This shows that the high similarity in terms of word class preference for the methods in Group 1 results in consistently lower agreement. Simultaneously, the similar preference for methods in Group 2, which happens to be close to human preference, correlates with higher agreement. From the opposite perspective: methods that are similar in terms of agreement scores exhibit similar word class preferences.

\subsection{Span Definition}\label{sec:span_definition}
We obtain syntactic spans by shallow parsing the data with Flair chunker \citep{akbik2018coling}, similarly to \citet{zhou-etal-2020-parsing} who use parsed constituents as pre-processed spans for a parsing experiment. Chunking is commonly adopted in Named Entity Recognition where usually noun phrases or verb phrases are the focus of interest \citep{taufiq2023named}. For our task, the advantage of this method over full constituency parsing \citep[e.g.]{kitaev-etal-2019-multilingual} or dependency parsing \citep[e.g.]{chen2014fast} is that the chunker output of discrete non-overlapping units facilitates direct alignment with attribution values. Punctuation tokens are ignored by the parser; we treat them as separate spans.

\citet{sikdar2021integrated} use constituency parsing  \citep{mrini2020rethinking} as a basis for hierarchically attributing feature importance scores from tokens to phrases (including any subphrases). However, different methods can have different word class preferences (e.g.\ a noun modifier may systematically be attributed more importance over its head) and it is therefore questionable whether score aggregation of any kind is a sensible approach. 
Having clearly defined, non-overlapping \textit{phrases} is instead crucial to our initial hypothesis.



In our dataset, each sentence contains on average 24.4 tokens (6--73), which are grouped into 15.3 spans (3--45). The average ratio of spans over tokens is 0.63 (0.23--1.0). A targeted span is a span that contains at least one token included in the top-$k$ selection by the attribution method. During agreement evaluation we treat spans as atomic units, meaning that a span is assigned 1 if targeted, otherwise 0 (similarly to tokens in top-$k$ selection). For a fixed $k$ set to 4, the average number of targeted spans in a sentence is slightly lower: Partition SHAP 3.5, LIME 3.6, Vanilla Grad 3.5, Grad\,×\,Input 3.6, Integrated Gradient 3.7, Integrated Gradient\,×\,Input 3.5, Human 3.3. The average over methods is 3.5.

\subsection{Head vs. Modifier Preference}
We have seen that Gradient\,×\,Input and Vanilla Gradient exhibit complementary linguistic preferences for noun tokens (the lowest versus highest ratio of noun tokens in the top-4). We zoom in on this phenomenon and investigate the attribution patterns in noun phrases (NPs), focussing on methods that select the head over its modifier and vice versa.

We examine a subset of noun phrase spans that are grouped according to $k=4$ by Gradient\,×\,Input and Vanilla Gradient. The NPs must span a minimum of two tokens to make the preference analysis for different word classes possible. To add some consensus stability to this subset, the spans under question should also be targeted by highly agreeing methods Partition SHAP and LIME. We compare the attribution profiles of Gradient\,×\,Input and Vanilla Gradient on the token and the span level for the specific [\textsc{det}, \textsc{noun}] construction, the most prevalent among length-2 noun phrases (73\%, 1,963). Interestingly, of the cases where Vanilla Gradient targeted \textsc{noun} (99\%, 1,951), Gradient\,×\,Input targeted \textsc{det} half of the times (899). This example clearly illustrates how methods do not only target different word classes in absolute terms, but also how that translates to systematic, alternating differences within syntactic spans.



Furthermore, the ratio of targeted tokens in the [\textsc{det}, \textsc{noun}] NPs is comparable: 57\% for Vanilla Gradient versus 60\% for Gradient\,×\,Input. This detail strengthens the claim of systematic preference in that the \textsc{det}--\textsc{noun} alternation, i.a., is usually \textit{exclusive}. In other words, it is uncommon for the two described methods to target both tokens from the NPs. This increases the prominence of the preference phenomenon in cases where one selects the \textsc{det} and the other the \textsc{noun}.

\section{Agreement at the Span Level}\label{sec:agreement_at_the_span_level}
We showed that different methods have different word class preferences and that the preference can be strong in the case of syntactic noun phrases. A consistently strong preference by two methods leads to a strong disagreement at the token level. The expectation that methods should agree on the token level might therefore be too strict. Given these insights, we measure method--method and human--method agreement at the span level, expecting a relative improvement compared to token-level agreement.

\subsection{Setup}
The dataset, model configurations and pool of attribution methods that we use are identical to those described in the linguistic analysis (§\ref{sec:Linguistic_analysis}). In addition, we adopt the definition for spans given in §\ref{sec:span_definition}. Our data therefore has a version where the instances are divided into tokens and one where instances are split into spans. The details of dynamic $k$ correspond to those described in §\ref{sec:topk_estimation}.


\subsection{The Effect of Dynamic $k$ on Spans}\label{sec:effectdynkonspans}
We compare the effect of dynamic $k$ on the span level versus dynamic $k$ on the token level. We measure the \textit{effect} as the increase in agreement i) versus a baseline to assess overall difficulty of the task and ii) versus fixed $k=4$ to assess the ability of the dynamic approach to detect important spans. We expect dynamic $k$ to be better suited than fixed $k$ to identify linguistic spans that the model considers important in the instance. Specifically, the \textit{local importance} setting (in combination with \textit{global importance}) appears to work as a pooling operator, highlighting the distinct important parts of the instance rather than few concentrated parts. We assume here that for the specific NLI task, $>1$ parts of the input should be considered important.

Agreement is measured as follows. We denote an attribution method as $\mathbf{A}$. $\mathbf{A}$ assigns an attribution profile $\mathbf{a}=\{a_1,a_2,...,a_n\}$ to the input sequence of tokens $\mathbf{s}=\{w_1,w_2,...,w_n\}$ so that each $a_i$ indicates the importance of token $w_i$ towards the inferred class. The subset of $k$ tokens with the highest attribution values are formalized as $\textrm{\textit{topk}}_A=\{t_1,t_2,...,t_k\}$. We compare $m$ attribution methods $A_1,...,A_m$ in pairs by calculating sentence-level agreement@$k$. Agreement@$k$ is based on the relevance of a each token. Relevance for a token $w_i$ is equal to the ratio of methods that include the token in their respective $topk$ subsets. Agreement@$k$ ignores perfect agreement on non-important tokens (where relevance $=0$) in order not to inflate the score. For our experiments, we report mean agreement@$k$, the averaged agreement over instances in the dataset $\mathbf{D} = \{s_1, s_2, \ldots, s_d\}$.

\vspace{-0.3cm}
\begin{alignat}{2}
&\textrm{Relevance }r(w_i) &&= 
\frac{\sum\limits_{A_j=1}^m{[ w_i \in \textrm{ \textit{topk}}_{A_j}]}
}{m}\\
&\textrm{Agreement}@k(s_i) &&= \frac{\sum\limits_{w_i=1}^n {r(w_i)}}{\sum\limits_{w_i=1}^n{[r(w_i)>0]}}
\end{alignat}


\vspace{-0.3cm}
\begin{alignat}{1}
\textrm{Agreement}@k(D) &= \frac{\sum\limits_{s_i=1}^d {\textrm{agreement}@k(s_i)}
}{d}
\end{alignat}

The average pair-wise agreement (all method--method combinations) for dynamic $k$ is 0.61 on the token level and 0.69 on the span level. 0.5 indicates perfect disagreement and 1.0 perfect agreement. While agreement seems relatively low, it might still suggest that consistently the same, few types of signals are identified by a pair of methods. 

We compute a baseline to measure how likely methods are to agree on the token and span level with a pseudo-random attribution method. In other words, we measure task difficulty of making two vectors with a subset of important tokens and spans to agree given a low value of $k$. For fixed $k=4$, 16\% of the tokens in a sentence would be highlighted on average; consequently, 23\% of the spans would be highlighted on average. For the token-level baseline, we then randomly shuffle two binary vectors of 100 elements, 16 of which 1s, and compute pairwise agreement. We repeat the process $10^3$ times. For the span-level baseline we adopt the same procedure, with the exception that the 1s in the vector are 23. The resulting baselines are 0.54 and 0.57, respectively, indicating that agreeing on tokens and spans is similarly difficult at low values of $k$. We thus observe the token-wise baseline for fixed $k$ being outperformed by 0.07 (0.54$\rightarrow$0.61), whereas the span-wise baseline is outperformed by a  relatively larger increase of 0.12 (0.57$\rightarrow$0.69).

\begin{figure}
\centering
\begin{subfigure}[b]{\linewidth}
  \centering
  \includegraphics[width=1.0\linewidth]{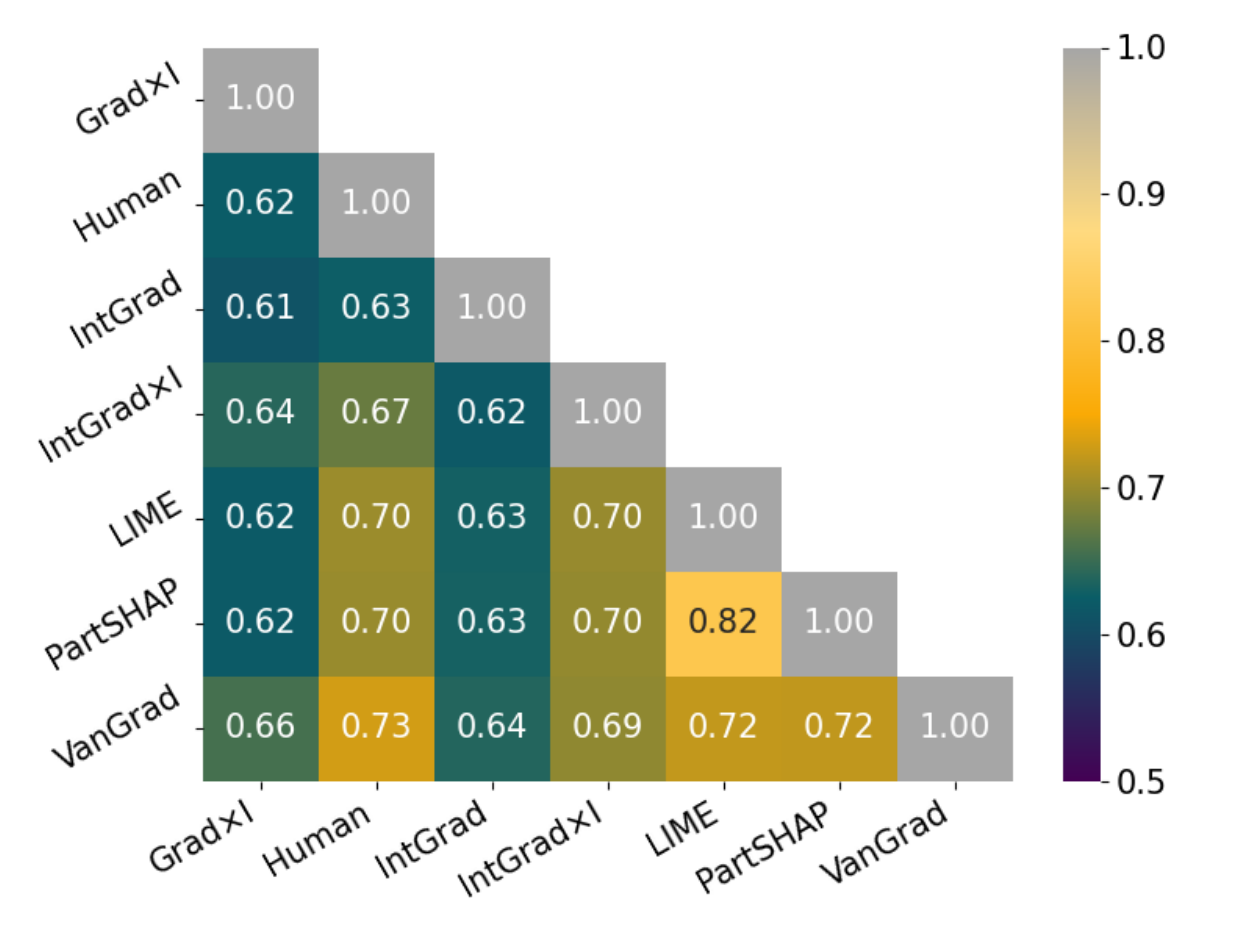}
  \caption{Mean span agreement@$k=4$.}
  \label{fig:pairwise_chunk_keq4}
\end{subfigure}
\begin{subfigure}[b]{\linewidth}
  \centering
  \includegraphics[width=1.0\linewidth]{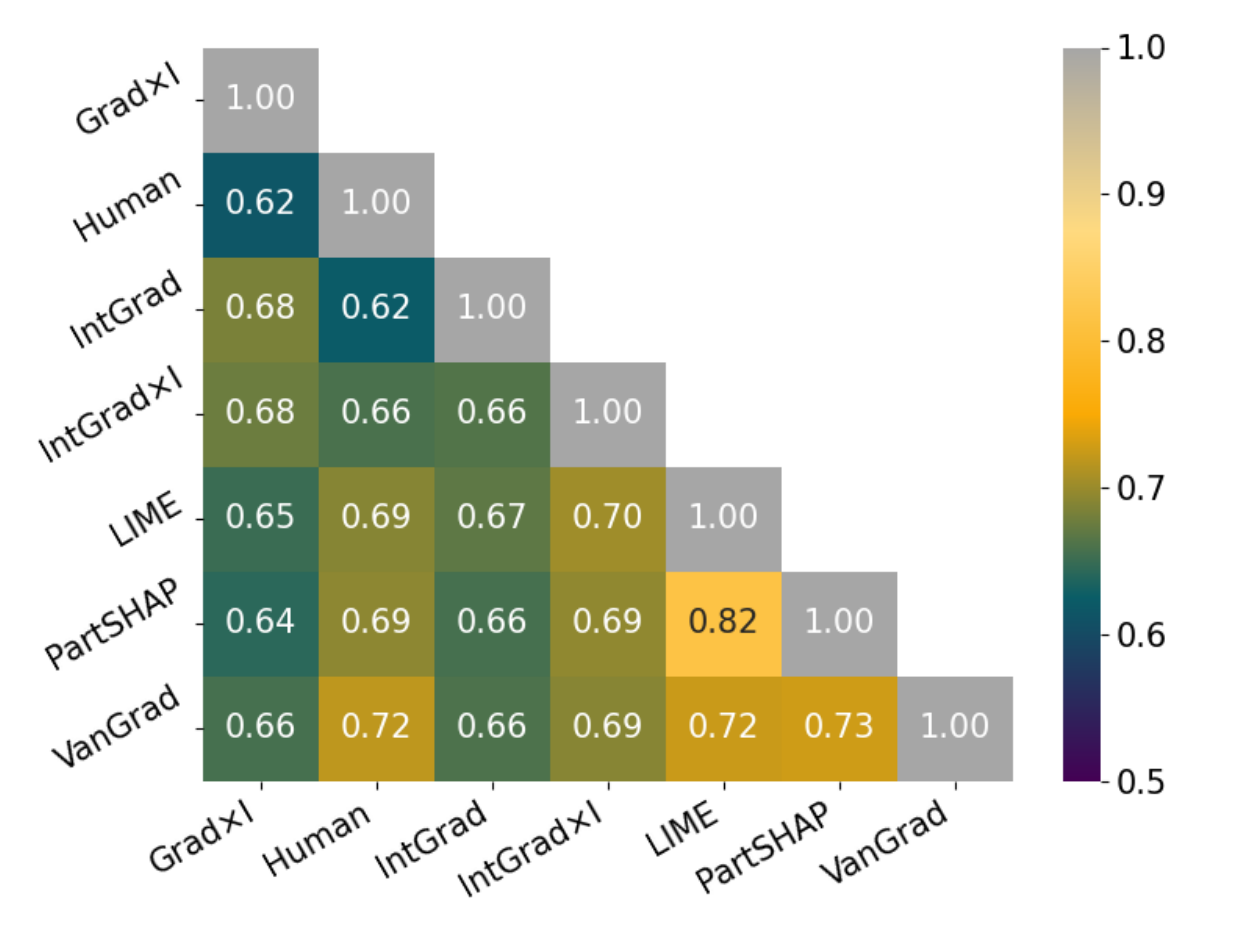}
  \caption{Mean span agreement@$k=$ dynamic.}
  \label{fig:pairwise_chunk_keqdyn}
\end{subfigure}
\caption{Span agreement for fixed and dynamic $k$.}
\label{fig:pairwise_chunk_fixedvsdyn}
\end{figure}

The comparison results between span agreement on fixed $k$ versus dynamic $k$ are given in Figure \ref{fig:pairwise_chunk_fixedvsdyn}. While dynamic $k$ provided marginal boosts (+0.00, +0.01, +0.02 compared to fixed $k=4$) on the token level for Gradient\,×\,Input and Integrated Gradient (compare \citet{kamp-etal-2023-dynamic}), it proves to have a larger positive effect on the span level. Specifically, the span agreement for method--method pairs that include Gradient\,×\,Input and/or Integrated Gradient remains constant or increases (changes from +0.00 to +0.07 compared to fixed $k$). At the same time, other method--method and human--method agreement scores remain constant or marginally decrease (+0.01, +0.00, -0.01). 

With regards to the largest difference observed between dynamic and fixed $k$, namely Integrated Gradient versus Gradient\,×\,Input, this can also be explained through the concentration levels of targeted tokens within spans. In fact, dynamic $k$ scatters the important tokens so that more spans are targeted compared to selecting an average fixed $k$. While $k=4$ yields 3.7 and 3.6 spans on average for the two methods, dynamic $k$ yields 6.9 and 6.5. Since it becomes easier for methods to agree when more tokens (and therefore more spans) are targeted, we investigate the settings of dynamic $k$ further (§\ref{sec:differentdynksettings}).

\subsection{Adjusting Dynamic $k$}\label{sec:differentdynksettings}
How can we validate or improve the dynamic $k$ algorithm? A solid global importance threshold should meet two conditions: i) resulting values of $k$ should be \textit{low}, preferably close to human preference average of 4$\pm$3; ii) they should outperform a baseline.

\begin{table*}[htbp]
    \centering
    \begin{tabular}{cccccccc}
        \toprule
        \multicolumn{1}{c}{} & & & & \textbf{Thresholds} & & & \\
        \midrule
        \multicolumn{1}{c}{\textbf{}} & \textbf{Method} & $\mu$ & $\mu + \sigma$ & $\mu + 2\sigma$ & $\mu - \sigma$ & $\mu - 2\sigma$ & \textit{median} \\
        \midrule
         & PartSHAP & \cellcolor{lightgray}4.54$\pm$1.73 & 2.16$\pm$0.95 & 1.25$\pm$0.65 & 7.36$\pm$ 1.89 & 7.37$\pm$1.89 & 6.19$\pm$1.62 \\
         & LIME & \cellcolor{lightgray}5.34$\pm$2.35 & 2.24$\pm$1.05 & 1.23$\pm$0.65 & 8.31$\pm$2.86 & 8.32$\pm$2.87 & 7.13$\pm$2.48 \\
         & VanGrad & \cellcolor{lightgray}4.58$\pm$1.68 & 2.41$\pm$1.02 & 1.39$\pm$0.61 & 7.63$\pm$2.68 & 7.64$\pm$2.69 & 6.20$\pm$2.08 \\
        all & Grad×I & \cellcolor{lightgray}6.83$\pm$2.59 & 2.39$\pm$1.12 & 0.68$\pm$0.65 & 8.21$\pm$2.82 & 8.28$\pm$2.83 & 7.08$\pm$2.51 \\
         & IntGrad & \cellcolor{lightgray}7.30$\pm$2.63 & 2.66$\pm$1.23 & 0.64$\pm$0.63 & 8.41$\pm$2.88 & 8.46$\pm$2.9 & 7.41$\pm$2.58 \\
         & IntGrad×I & \cellcolor{lightgray}5.68$\pm$2.37 & 2.27$\pm$1.08 & 1.02$\pm$0.62 & 8.04$\pm$2.80 & 8.07$\pm$2.82 & 6.83$\pm$2.39 \\
        \midrule
        \midrule
         & PartSHAP & \cellcolor{lightgray}3.34$\pm$1.33 & 1.86$\pm$0.82 & 1.07$\pm$0.55 & 7.00$\pm$2.01 & 7.28$\pm$1.93 & \cellcolor{lightgray}5.01$\pm$1.61 \\
         & LIME & \cellcolor{lightgray}3.56$\pm$1.56 & 1.87$\pm$0.87 & 1.06$\pm$0.53 & 7.95$\pm$2.91 & 8.25$\pm$2.89 & \cellcolor{lightgray}5.59$\pm$2.04 \\
         & VanGrad & \cellcolor{lightgray}4.58$\pm$1.68 & 2.41$\pm$1.02 & 1.39$\pm$0.61 & 7.63$\pm$2.68 & 7.64$\pm$2.69 & \cellcolor{lightgray}6.20$\pm$2.08 \\
        $>0$ & Grad×I & \cellcolor{lightgray}3.51$\pm$1.56 & 1.75$\pm$0.83 & 0.73$\pm$0.56 & 6.81$\pm$2.82 & 7.86$\pm$2.92 & \cellcolor{lightgray}4.69$\pm$1.88 \\
         & IntGrad & \cellcolor{lightgray}3.47$\pm$1.60 & 1.67$\pm$0.81 & 0.62$\pm$0.55 & 6.60$\pm$2.81 & 7.81$\pm$3.05 & \cellcolor{lightgray}4.54$\pm$1.86 \\
         & IntGrad×I & \cellcolor{lightgray}3.83$\pm$1.69 & 1.91$\pm$0.90 & 0.98$\pm$0.53 & 7.57$\pm$2.74 & 7.99$\pm$2.81 & \cellcolor{lightgray}5.38$\pm$1.96 \\
        \bottomrule
    \end{tabular}
    \caption{Values of \textit{k} for different global importance thresholds. The three methods that yield values of $k$ closest to human preference are visually indicated with a dark background.}
    \label{tab:thresholds}
\end{table*}

We explore multiple thresholds: different combinations of $\mu [+,-] [0,1,2] \sigma$, typical distances from the mean in a distribution; the \textit{median}, which is more robust to outliers than $\mu$. The thresholds are calculated for (a) all scores and (b) positive scores. Thresholds for positive scores should ignore attributions with negative importance towards the inferred class. These are common in methods such as Integrated Gradient\,×\,Input. The influence of negative values and peaks in the attribution profiles is not accounted for by the current threshold set at $\mu$.

The resulting values of $k$ for different thresholds are given in Table \ref{tab:thresholds}. We find that for different thresholds, resulting $k$s are comparable across methods, which might indicate that the attribution profiles have overall similar distributions. The three thresholds that yield closest $k$s to human preference are $\mu$, $\mu>0$ and \textit{median} $>0$. Closeness corresponds to the averaged Euclidean distance between the mean$\pm$stdev pairs and human preference of 4$\pm$3, for each threshold column.\footnote{See Appendix \ref{appendix} for an overview of the distances.} Among these three, $\mu$ had already proven to keep $k$ low and close to ground truth average \citep{kamp-etal-2023-dynamic}.

Even if the estimated $k$s by the three candidate thresholds are relatively low, it could be, for example, that a method-specific $k$ is too high, positively biasing the agreement score. A high $k$ would even give high agreement for a pseudo-random attribution profile, which should not be possible if the threshold is properly set. Hence, we compare each method's agreement scores with other methods to the method's agreement with a baseline. This gives us an indication of how well a specific threshold works with different attribution profiles. We do this both on the token level and on the span level. The baseline method operates pseudo-randomly by assigning attribution scores to the tokens without knowledge about token importance. For each method, we randomly shuffle the scores in each attribution profile. Each method has its own baseline so that the different distributional properties of the attribution profiles are preserved. We then compute agreement@dynamic-$k$ between original and shuffled attribution profiles, which are consequently averaged over the dataset. If the threshold for $k$-estimation is strong, the agreement with the baseline for each method should be lower than the agreement with other methods. 

\begin{table}[h]
\centering
\begin{tabular}{ccc}
\toprule
& \textbf{Token} & \textbf{Span} \\
\midrule
\textbf{Method} & \multicolumn{2}{c}{\textit{BL:minAgr--maxAgr}} \\
\midrule
PartSHAP & 0.56:0.56--0.78 & 0.64:0.64--0.82 \\
LIME & 0.57:0.57--0.78 & 0.65:0.65--0.82 \\
VanGrad & 0.56:0.58--0.68 & 0.64:0.66--0.73 \\
Grad×I & 0.59:\textbf{0.56}--0.60 & 0.68:\textbf{0.64}--0.69 \\
IntGrad & 0.60:\textbf{0.58}--\textbf{0.59} & 0.69:\textbf{0.66}--\textbf{0.68} \\
IntGrad×I & 0.58:0.58--0.64 & 0.66:0.66--0.70 \\
\bottomrule
\end{tabular}
\caption{Token and span agr. with other methods (range \textit{minAgr} to \textit{maxAgr}) versus baseline (\textit{BL}), for threshold $=\mu$. Scores $<$ baseline in \textbf{bold}.}
\label{table:baseline_token_span_mu}
\end{table}

Results for $\mu$ are given in Table \ref{table:baseline_token_span_mu}. We find that for $\mu$, Integrated Gradient and Gradient\,×\,Input have higher baseline agreement than the other methods. This can be explained by the higher values of $k$ for this threshold (i.e.\ 6.83 and 7.30 in Table \ref{tab:thresholds}). Importantly, both methods have method--method agreement scores that do not beat the baseline (which pseudo-randomly selects tokens), neither on the token level nor on the span level. With regards to \textit{median} $>0$,\footnote{While reporting the baseline tests for threshold $=\mu$ in Table \ref{table:baseline_token_span_mu}, we leave the overviews for thresholds $\mu>0$ and \textit{median} $>0$ to Appendix \ref{appendix}.} multiple methods do not beat their baselines either. The threshold $\mu>0$ instead does, for all methods and both on tokens and on spans. This is an indication of the fact that the latter might be a better threshold than $\mu$ for dynamic $k$ estimation. An additional interpretation of why $\mu>0$ works better than $\mu$ is that negative local maxima in the attribution profiles are hereby ignored, leading to less but more important $k$ tokens (and spans) to be targeted. This baseline testing also shows that Gradient\,×\,Input and Integrated Gradient are unreliable methods: they have low agreement with other methods and often fail to beat a random baseline. 


\section{Discussion}
Analysing disagreement from a linguistic perspective helps us to better understand the differences between attribution methods. We briefly discuss the implications of token- and span-level analyses on other tasks than NLI. With an eye on the ability and reliability of these methods to reflect the model's decision process, we also consider the implications for the \textit{faithfulness} aspect in interpretability research.

\paragraph{Generalisability of Spans}
Generally speaking, an NLI task is sufficiently challenging that it avoids sentences of different classes (e.g.\ contradiction, entailment) differing by exactly one word. It is therefore fair to expect methods to target the same span and not to penalise them for disagreeing on the token level. However, targeting a modifier instead of its syntactic head can make a big difference for other tasks. Additionally, the span-token ratio should determine the difficulty of assessing span-level agreement compared to tokens. 
The choice of considering spans rather than tokens should therefore be weighted against the type of task and data. 

On a similar note, §\ref{sec:token_type_preference} describes the systematic differences in punctuation preferences. We may hypothesise that methods that consistently include full stops in their top-$k$ are actually catching the signal's onset (\textit{locality} information) rather than the full stop being itself a signal (\textit{lexical} information). To this end, our choice of treating punctuation as separate spans might have influenced the span agreement of such methods. More research is necessary to disentangle locality from lexical information.

\paragraph{Agreement as a Proxy for Faithfulness}
Agreement is linked with both plausibility and faithfulness. We considered plausibility when estimating dynamic $k$ thresholds, as we aimed for $k$s close to human preference. However, a more direct way of testing for plausibility in this context is by assessing human--method agreement, which we mostly left out of scope in this study. To that end, we did find that agreement results are constant on both tokens and spans, possibly suggesting that human--method agreement reaches a ceiling already at the token level (i.e.\ tokens are targeted that belong to different signals in the sentence). This interpretation might even hold for more faithful methods. In fact, models do often not rely on the same patterns as humans do, instead resorting to shortcut signals.

Measuring faithfulness, on the other hand, is less straightforward. Following \citet{jain2019attention}, who state that faithful attention-based explanations should be agreeable, we carefully extend their perspective in that agreement between method-\textit{generic} explanations can be considered as a proxy for faithfulness. According to the principle of reproducibility in science \citep{popper2005logic}, a finding that is confirmed through different means is, in principle, more likely to be correct. As such, if two attribution methods with distinct means yield similar results, they are likely similarly (un)faithful. If one method disagrees with the majority of the batch, either the one, the majority, or all are unfaithful. Because of the reproducibility principle, however, it is more likely that the majority is more faithful.

In this light, we could therefore speculate that Gradient\,×\,Input and Integrated Gradient were two of the less faithful methods in our study, an argument that is supported by their scarce agreement compared to a pseudo-random baseline. Given that some methods might highly correlate with other methods by design, one must be careful at drawing conclusions. Constructing a batch of methods that is representative of different ways of interpreting the model is, for this reason, not a simple task.

\section{Conclusion and Future Directions}
In this study, we approached post-hoc explanation disagreement from a syntactic perspective. We found that methods that agree most with other methods and with aggregated scores of human rationales have similar POS tag preferences for the targeted tokens. We then determined that attribution methods agree more at the span level than at the token level, which appear to be similarly difficult tasks at low values of $k$. One particular reason for disagreement is the consistent preference by one method to target the determiners instead of the noun head within the same noun phrase. We showed that dynamic $k$ works well in combination with spans, as it seeks for non-neighboring important signals in the sentence. Finally, we empirically tested for different thresholds of the \textit{global importance} setting of dynamic $k$, suggesting a value ($\mu>0$) that accounts for both negative attribution scores and results in low $k$s.

One issue that dynamic $k$ aims to tackle is the targeting of redundant tokens as signals in the same span. To complement this, a more in-depth analysis would provide a better understanding about the way that different methods concentrate their targeted tokens in the same spans. Intuitively, for a fixed $k$, some methods highlight tokens that are more sparse across the instance, whereas other more quickly concentrate targeted tokens within the same spans. To obtain such a concentration metric, one could measure how rapidly a set of tokens belonging to the most important ground truth span are being targeted, at increasing values of $k$. 

Future directions of research include the exploration of different \textit{local importance} criteria in the dynamic $k$ algorithm, such as different windows (current $\pm$1 versus $\pm$2, $\pm$3). Another is to exploit (syntactic) span-based information to improve interpretability accuracy at the token level, or to improve explanation aggregation techniques. Finally, we advise future evaluation datasets based on multiple annotators' rationales to preserve specific instance--annotator mappings in the metadata. This would facilitate new directions in assessing the plausibility of attribution methods, specifically how variations in human subjectivity relate to agreement.

\section{Ethical Considerations}
We would like to reiterate that attribution scores cannot be blindly relied upon to precisely determine model functioning, as they can be influenced by experimental factors such as task and model performance. To avoid drawing generalised conclusions, it is advisable to employ multiple metrics when studying feature attribution.

\section{Acknowledgements}
Jonathan Kamp’s research was funded by the Dutch National Science Organisation (NWO) through the project InDeep: Interpreting Deep Learning Models for Text and Sound (NWA.1292.19.399). Antske Fokkens was supported by the EU Horizon 2020 project InTaVia: In/Tangible European Heritage - Visual Analysis, Curation and Communication (http://intavia.eu) under grant agreement No. 101004825. Lisa Beinborn's work was funded by the Dutch National Science Organisation (NWO) through the VENI program (Vl.Veni.211C.039). We would like to thank the anonymous reviewers for their valuable contribution.

\section{Bibliographical References}\label{sec:reference}

\bibliographystyle{lrec-coling2024-natbib}
\bibliography{lrec-coling2024}

\begin{thebibliography}{1}
\expandafter\ifx\csname natexlab\endcsname\relax\def\natexlab#1{#1}\fi

\bibitem[{Camburu et~al.(2018)Camburu, Rockt{\"a}schel, Lukasiewicz, and Blunsom}]{camburu2018snli}
Camburu, Oana-Maria and Rockt{\"a}schel, Tim and Lukasiewicz, Thomas and Blunsom, Phil. 2018.
\newblock \emph{e-snli: Natural language inference with natural language explanations}.
\newblock GitHub Repository without PID/islrn: https://github.com/OanaMariaCamburu/e-SNLI.

\end{thebibliography}


\begin{thebibliography}{40}
\expandafter\ifx\csname natexlab\endcsname\relax\def\natexlab#1{#1}\fi

\bibitem[{Akbik et~al.(2018)Akbik, Blythe, and Vollgraf}]{akbik2018coling}
Alan Akbik, Duncan Blythe, and Roland Vollgraf. 2018.
\newblock Contextual string embeddings for sequence labeling.
\newblock In \emph{{COLING} 2018, 27th International Conference on Computational Linguistics}, pages 1638--1649.

\bibitem[{Atanasova et~al.(2023)Atanasova, Camburu, Lioma, Lukasiewicz, Simonsen, and Augenstein}]{atanasova-etal-2023-faithfulness}
Pepa Atanasova, Oana-Maria Camburu, Christina Lioma, Thomas Lukasiewicz, Jakob~Grue Simonsen, and Isabelle Augenstein. 2023.
\newblock \href {https://doi.org/10.18653/v1/2023.acl-short.25} {Faithfulness tests for natural language explanations}.
\newblock In \emph{Proceedings of the 61st Annual Meeting of the Association for Computational Linguistics (Volume 2: Short Papers)}, pages 283--294, Toronto, Canada. Association for Computational Linguistics.

\bibitem[{Atanasova et~al.(2020)Atanasova, Simonsen, Lioma, and Augenstein}]{atanasova2020diagnostic}
Pepa Atanasova, Jakob~Grue Simonsen, Christina Lioma, and Isabelle Augenstein. 2020.
\newblock A diagnostic study of explainability techniques for text classification.
\newblock In \emph{Proceedings of the 2020 Conference on Empirical Methods in Natural Language Processing (EMNLP)}, pages 3256--3274.

\bibitem[{Attanasio et~al.(2022)Attanasio, Nozza, Pastor, Hovy et~al.}]{attanasio2022benchmarking}
Giuseppe Attanasio, Debora Nozza, Eliana Pastor, Dirk Hovy, et~al. 2022.
\newblock Benchmarking post-hoc interpretability approaches for transformer-based misogyny detection.
\newblock In \emph{Proceedings of NLP Power! The First Workshop on Efficient Benchmarking in NLP}. Association for Computational Linguistics.

\bibitem[{Attanasio et~al.(2023)Attanasio, Pastor, Di~Bonaventura, and Nozza}]{attanasio-etal-2023-ferret}
Giuseppe Attanasio, Eliana Pastor, Chiara Di~Bonaventura, and Debora Nozza. 2023.
\newblock \href {https://doi.org/10.18653/v1/2023.eacl-demo.29} {ferret: a framework for benchmarking explainers on transformers}.
\newblock In \emph{Proceedings of the 17th Conference of the European Chapter of the Association for Computational Linguistics: System Demonstrations}, pages 256--266, Dubrovnik, Croatia. Association for Computational Linguistics.

\bibitem[{Babiker et~al.(2023)Babiker, Kim, and Goebel}]{babiker2023intermediate}
Housam~KB Babiker, Mi-Young Kim, and Randy Goebel. 2023.
\newblock From intermediate representations to explanations: Exploring hierarchical structures in nlp.
\newblock In \emph{ECAI 2023}, pages 157--164. IOS Press.

\bibitem[{Bastings et~al.(2022)Bastings, Ebert, Zablotskaia, Sandholm, and Filippova}]{bastings-etal-2022-will}
Jasmijn Bastings, Sebastian Ebert, Polina Zablotskaia, Anders Sandholm, and Katja Filippova. 2022.
\newblock \href {https://doi.org/10.18653/v1/2022.emnlp-main.64} {{``}will you find these shortcuts?{''} a protocol for evaluating the faithfulness of input salience methods for text classification}.
\newblock In \emph{Proceedings of the 2022 Conference on Empirical Methods in Natural Language Processing}, pages 976--991, Abu Dhabi, United Arab Emirates. Association for Computational Linguistics.

\bibitem[{Bongard et~al.(2022)Bongard, Held, and Habernal}]{bongard2022legal}
Leonard Bongard, Lena Held, and Ivan Habernal. 2022.
\newblock The legal argument reasoning task in civil procedure.
\newblock In \emph{Proceedings of the Natural Legal Language Processing Workshop 2022}, pages 194--207.

\bibitem[{Camburu et~al.(2019)Camburu, Giunchiglia, Foerster, Lukasiewicz, and Blunsom}]{camburu2019can}
Oana-Maria Camburu, Eleonora Giunchiglia, Jakob Foerster, Thomas Lukasiewicz, and Phil Blunsom. 2019.
\newblock Can i trust the explainer? verifying post-hoc explanatory methods.
\newblock \emph{arXiv preprint arXiv:1910.02065}.

\bibitem[{Carton et~al.(2020)Carton, Rathore, and Tan}]{carton-etal-2020-evaluating}
Samuel Carton, Anirudh Rathore, and Chenhao Tan. 2020.
\newblock \href {https://doi.org/10.18653/v1/2020.emnlp-main.747} {Evaluating and characterizing human rationales}.
\newblock In \emph{Proceedings of the 2020 Conference on Empirical Methods in Natural Language Processing (EMNLP)}, pages 9294--9307, Online. Association for Computational Linguistics.

\bibitem[{Chen and Manning(2014)}]{chen2014fast}
Danqi Chen and Christopher~D Manning. 2014.
\newblock A fast and accurate dependency parser using neural networks.
\newblock In \emph{Proceedings of the 2014 conference on empirical methods in natural language processing (EMNLP)}, pages 740--750.

\bibitem[{Choudhary et~al.(2022)Choudhary, Chatterjee, and Saha}]{choudhary2022interpretation}
Shivani Choudhary, Niladri Chatterjee, and Subir~Kumar Saha. 2022.
\newblock Interpretation of black box nlp models: A survey.
\newblock \emph{arXiv preprint arXiv:2203.17081}.

\bibitem[{Devlin et~al.(2019)Devlin, Chang, Lee, and Toutanova}]{devlin-etal-2019-bert}
Jacob Devlin, Ming-Wei Chang, Kenton Lee, and Kristina Toutanova. 2019.
\newblock \href {https://doi.org/10.18653/v1/N19-1423} {{BERT}: Pre-training of deep bidirectional transformers for language understanding}.
\newblock In \emph{Proceedings of the 2019 Conference of the North {A}merican Chapter of the Association for Computational Linguistics: Human Language Technologies, Volume 1 (Long and Short Papers)}, pages 4171--4186, Minneapolis, Minnesota. Association for Computational Linguistics.

\bibitem[{Jacovi and Goldberg(2020)}]{jacovi2020towards}
Alon Jacovi and Yoav Goldberg. 2020.
\newblock Towards faithfully interpretable nlp systems: How should we define and evaluate faithfulness?
\newblock In \emph{Proceedings of the 58th Annual Meeting of the Association for Computational Linguistics}, pages 4198--4205.

\bibitem[{Jain and Wallace(2019)}]{jain2019attention}
Sarthak Jain and Byron~C Wallace. 2019.
\newblock Attention is not explanation.
\newblock In \emph{Proceedings of the 2019 Conference of the North American Chapter of the Association for Computational Linguistics: Human Language Technologies, Volume 1 (Long and Short Papers)}, pages 3543--3556.

\bibitem[{Jesus et~al.(2021)Jesus, Bel{\'e}m, Balayan, Bento, Saleiro, Bizarro, and Gama}]{jesus2021can}
S{\'e}rgio Jesus, Catarina Bel{\'e}m, Vladimir Balayan, Jo{\~a}o Bento, Pedro Saleiro, Pedro Bizarro, and Jo{\~a}o Gama. 2021.
\newblock How can i choose an explainer? an application-grounded evaluation of post-hoc explanations.
\newblock In \emph{Proceedings of the 2021 ACM Conference on Fairness, Accountability, and Transparency}, pages 805--815.

\bibitem[{Jumelet and Zuidema(2023)}]{jumelet-zuidema-2023-feature}
Jaap Jumelet and Willem Zuidema. 2023.
\newblock \href {https://doi.org/10.18653/v1/2023.findings-acl.554} {Feature interactions reveal linguistic structure in language models}.
\newblock In \emph{Findings of the Association for Computational Linguistics: ACL 2023}, pages 8697--8712, Toronto, Canada. Association for Computational Linguistics.

\bibitem[{Kamp et~al.(2023)Kamp, Beinborn, and Fokkens}]{kamp-etal-2023-dynamic}
Jonathan Kamp, Lisa Beinborn, and Antske Fokkens. 2023.
\newblock \href {https://doi.org/10.18653/v1/2023.emnlp-main.379} {Dynamic top-k estimation consolidates disagreement between feature attribution methods}.
\newblock In \emph{Proceedings of the 2023 Conference on Empirical Methods in Natural Language Processing}, pages 6190--6197, Singapore. Association for Computational Linguistics.

\bibitem[{Kitaev et~al.(2019)Kitaev, Cao, and Klein}]{kitaev-etal-2019-multilingual}
Nikita Kitaev, Steven Cao, and Dan Klein. 2019.
\newblock \href {https://doi.org/10.18653/v1/P19-1340} {Multilingual constituency parsing with self-attention and pre-training}.
\newblock In \emph{Proceedings of the 57th Annual Meeting of the Association for Computational Linguistics}, pages 3499--3505, Florence, Italy. Association for Computational Linguistics.

\bibitem[{Krishna et~al.(2022)Krishna, Han, Gu, Pombra, Jabbari, Wu, and Lakkaraju}]{krishna2022disagreement}
Satyapriya Krishna, Tessa Han, Alex Gu, Javin Pombra, Shahin Jabbari, Steven Wu, and Himabindu Lakkaraju. 2022.
\newblock The disagreement problem in explainable machine learning: A practitioner's perspective.
\newblock \emph{arXiv preprint arXiv:2202.01602}.

\bibitem[{Lai et~al.(2019)Lai, Cai, and Tan}]{lai2019many}
Vivian Lai, Zheng Cai, and Chenhao Tan. 2019.
\newblock Many faces of feature importance: Comparing built-in and post-hoc feature importance in text classification.
\newblock In \emph{Proceedings of the 2019 Conference on Empirical Methods in Natural Language Processing and the 9th International Joint Conference on Natural Language Processing (EMNLP-IJCNLP)}, pages 486--495.

\bibitem[{Lundberg and Lee(2017)}]{lundberg2017unified}
Scott~M Lundberg and Su-In Lee. 2017.
\newblock A unified approach to interpreting model predictions.
\newblock \emph{Advances in neural information processing systems}, 30.

\bibitem[{Madsen et~al.(2022)Madsen, Reddy, and Chandar}]{madsen2022post}
Andreas Madsen, Siva Reddy, and Sarath Chandar. 2022.
\newblock Post-hoc interpretability for neural nlp: A survey.
\newblock \emph{ACM Computing Surveys}, 55(8):1--42.

\bibitem[{Mrini et~al.(2020)Mrini, Dernoncourt, Tran, Bui, Chang, and Nakashole}]{mrini2020rethinking}
Khalil Mrini, Franck Dernoncourt, Quan~Hung Tran, Trung Bui, Walter Chang, and Ndapandula Nakashole. 2020.
\newblock Rethinking self-attention: Towards interpretability in neural parsing.
\newblock In \emph{Findings of the Association for Computational Linguistics: EMNLP 2020}, pages 731--742.

\bibitem[{Neely et~al.(2022)Neely, Schouten, Bleeker, and Lucic}]{neely2022song}
Michael Neely, Stefan~F Schouten, Maurits Bleeker, and Ana Lucic. 2022.
\newblock A song of (dis) agreement: Evaluating the evaluation of explainable artificial intelligence in natural language processing.
\newblock In \emph{HHAI2022: Augmenting Human Intellect}, pages 60--78. IOS Press.

\bibitem[{Pirie et~al.(2023)Pirie, Wiratunga, Wijekoon, and Moreno-Garcia}]{pirie2023agree}
Craig Pirie, Nirmalie Wiratunga, Anjana Wijekoon, and Carlos~Francisco Moreno-Garcia. 2023.
\newblock {AGREE}: a feature attribution aggregation framework to address explainer disagreements with alignment metrics.
\newblock In \emph{Proceedings of the Workshops at the 31st International Conference on Case-Based Reasoning (ICCBR-WS 2023)}, pages 184--199. CEUR.

\bibitem[{Popper(2005)}]{popper2005logic}
Karl Popper. 2005.
\newblock \emph{The logic of scientific discovery}.
\newblock Routledge.

\bibitem[{Pruthi et~al.(2022)Pruthi, Bansal, Dhingra, Soares, Collins, Lipton, Neubig, and Cohen}]{pruthi2022evaluating}
Danish Pruthi, Rachit Bansal, Bhuwan Dhingra, Livio~Baldini Soares, Michael Collins, Zachary~C Lipton, Graham Neubig, and William Cohen. 2022.
\newblock Evaluating explanations: How much do explanations from the teacher aid students?
\newblock \emph{Transactions of the Association for Computational Linguistics}, 10:359--375.

\bibitem[{Ramnath et~al.(2020)Ramnath, Nema, Sahni, and Khapra}]{ramnath2020towards}
Sahana Ramnath, Preksha Nema, Deep Sahni, and Mitesh~M Khapra. 2020.
\newblock Towards interpreting bert for reading comprehension based qa.
\newblock In \emph{Proceedings of the 2020 Conference on Empirical Methods in Natural Language Processing (EMNLP)}, pages 3236--3242.

\bibitem[{R{\"a}uker et~al.(2023)R{\"a}uker, Ho, Casper, and Hadfield-Menell}]{rauker2023toward}
Tilman R{\"a}uker, Anson Ho, Stephen Casper, and Dylan Hadfield-Menell. 2023.
\newblock Toward transparent ai: A survey on interpreting the inner structures of deep neural networks.
\newblock In \emph{2023 IEEE Conference on Secure and Trustworthy Machine Learning (SaTML)}, pages 464--483. IEEE.

\bibitem[{Ribeiro et~al.(2016)Ribeiro, Singh, and Guestrin}]{ribeiro-etal-2016-trust}
Marco Ribeiro, Sameer Singh, and Carlos Guestrin. 2016.
\newblock \href {https://doi.org/10.18653/v1/N16-3020} {{``}why should {I} trust you?{''}: Explaining the predictions of any classifier}.
\newblock In \emph{Proceedings of the 2016 Conference of the North {A}merican Chapter of the Association for Computational Linguistics: Demonstrations}, pages 97--101, San Diego, California. Association for Computational Linguistics.

\bibitem[{Roy et~al.(2022)Roy, Laberge, Roy, Khomh, Nikanjam, and Mondal}]{roy2022don}
Saumendu Roy, Gabriel Laberge, Banani Roy, Foutse Khomh, Amin Nikanjam, and Saikat Mondal. 2022.
\newblock Why don’t xai techniques agree? characterizing the disagreements between post-hoc explanations of defect predictions.
\newblock In \emph{2022 IEEE International Conference on Software Maintenance and Evolution (ICSME)}, pages 444--448. IEEE.

\bibitem[{Sanh et~al.(2019)Sanh, Debut, Chaumond, and Wolf}]{sanh2019distilbert}
Victor Sanh, Lysandre Debut, Julien Chaumond, and Thomas Wolf. 2019.
\newblock Distilbert, a distilled version of bert: smaller, faster, cheaper and lighter.
\newblock \emph{arXiv preprint arXiv:1910.01108}.

\bibitem[{Shrikumar et~al.(2017)Shrikumar, Greenside, and Kundaje}]{shrikumar2017learning}
Avanti Shrikumar, Peyton Greenside, and Anshul Kundaje. 2017.
\newblock Learning important features through propagating activation differences.
\newblock In \emph{Proceedings of the 34th International Conference on Machine Learning - Volume 70}, ICML'17, page 3145–3153. JMLR.org.

\bibitem[{Sikdar et~al.(2021)Sikdar, Bhattacharya, and Heese}]{sikdar2021integrated}
Sandipan Sikdar, Parantapa Bhattacharya, and Kieran Heese. 2021.
\newblock Integrated directional gradients: Feature interaction attribution for neural nlp models.
\newblock In \emph{Proceedings of the 59th Annual Meeting of the Association for Computational Linguistics and the 11th International Joint Conference on Natural Language Processing (Volume 1: Long Papers)}, pages 865--878.

\bibitem[{Simonyan et~al.(2014)Simonyan, Vedaldi, and Zisserman}]{simonyan2014deep}
K~Simonyan, A~Vedaldi, and A~Zisserman. 2014.
\newblock Deep inside convolutional networks: visualising image classification models and saliency maps.
\newblock In \emph{Proceedings of the International Conference on Learning Representations (ICLR)}. ICLR.

\bibitem[{Song et~al.(2023)Song, Giunchiglia, Li, and Xu}]{song2023automatic}
Rui Song, Fausto Giunchiglia, Yingji Li, and Hao Xu. 2023.
\newblock Automatic counterfactual augmentation for robust text classification based on word-group search.
\newblock \emph{arXiv preprint arXiv:2307.01214}.

\bibitem[{Sundararajan et~al.(2017)Sundararajan, Taly, and Yan}]{sundararajan2017axiomatic}
Mukund Sundararajan, Ankur Taly, and Qiqi Yan. 2017.
\newblock Axiomatic attribution for deep networks.
\newblock In \emph{International conference on machine learning}, pages 3319--3328. PMLR.

\bibitem[{Taufiq et~al.(2023)Taufiq, Pulungan, and Suyanto}]{taufiq2023named}
Umar Taufiq, Reza Pulungan, and Yohanes Suyanto. 2023.
\newblock Named entity recognition and dependency parsing for better concept extraction in summary obfuscation detection.
\newblock \emph{Expert Systems with Applications}, 217:119579.

\bibitem[{Zhou et~al.(2020)Zhou, Li, and Zhao}]{zhou-etal-2020-parsing}
Junru Zhou, Zuchao Li, and Hai Zhao. 2020.
\newblock \href {https://doi.org/10.18653/v1/2020.findings-emnlp.398} {Parsing all: Syntax and semantics, dependencies and spans}.
\newblock In \emph{Findings of the Association for Computational Linguistics: EMNLP 2020}, pages 4438--4449, Online. Association for Computational Linguistics.

\end{thebibliography}

\section{Language Resource References}\label{lr:ref}

\bibliographystylelanguageresource{lrec-coling2024-natbib}
\bibliographylanguageresource{languageresource}

\appendix

\section{Appendix}\label{appendix}
The baseline tests for importance thresholds $\mu>0$ and \textit{median} $>0$ described in §\ref{sec:differentdynksettings} are given in Table \ref{table:baseline_token_span_mu>0} and Table \ref{table:baseline_token_span_median>0}, respectively. The averaged Euclidean distances that led to selecting these thresholds (§\ref{sec:differentdynksettings}) are reported in Table \ref{tab:euclidean_distances}. In Table \ref{tab:chi_squared_results}, we find the results of the Chi-Square tests adopted within the linguistic analysis in §\ref{sec:token_type_preference}.

\begin{table}[h]
\centering
\begin{tabular}{ccc}
\toprule
& \textbf{Token} & \textbf{Span} \\
\midrule
\textbf{Method} & \multicolumn{2}{c}{\textit{BL:minAgr--maxAgr}} \\
\midrule
PartSHAP & 0.55:0.55--0.81 & 0.60:0.61--0.83 \\
LIME & 0.55:0.55--0.81 & 0.60:0.61--0.83 \\
VanGrad & 0.56:0.58--0.68 & 0.64:0.64--0.72 \\
Grad×I & 0.55:0.55--0.58 & 0.60:0.60--0.65 \\
IntGrad & 0.55:0.55--0.58 & 0.59:0.61--0.64 \\
IntGrad×I & 0.55:0.57--0.65 & 0.61:0.61--0.69 \\
\bottomrule
\end{tabular}
\caption{Token and span agr. with other methods (range \textit{minAgr} to \textit{maxAgr}) versus baseline (\textit{BL}), for thresh. $=\mu>0$. Scores $<$ baseline in \textbf{bold}.}
\label{table:baseline_token_span_mu>0}
\end{table}

\begin{table}[h]
\centering
\begin{tabular}{ccc}
\toprule
& \textbf{Token} & \textbf{Span} \\
\midrule
\textbf{Method} & \multicolumn{2}{c}{\textit{BL:minAgr--maxAgr}} \\
\midrule
PartSHAP & 0.57:\textbf{0.56}--0.76 & 0.65:\textbf{0.63}--0.80 \\
LIME & 0.57:\textbf{0.56}--0.76 & 0.65:\textbf{0.63}--0.80 \\
VanGrad & 0.59:\textbf{0.58}--0.68 & 0.69:\textbf{0.66}--0.74 \\
Grad×I & 0.56:0.56--0.59 & 0.63:\textbf{0.62}--0.67 \\
IntGrad & 0.56:0.56--0.58 & 0.62:0.62--0.66 \\
IntGrad×I & 0.57:0.57--0.64 & 0.65:0.66--0.74 \\
\bottomrule
\end{tabular}
\caption{Token and span agr. with other methods (range \textit{minAgr} to \textit{maxAgr}) vs. baseline (\textit{BL}), for thresh. $=$ \textit{median} $>0$. Scores $<$ baseline in \textbf{bold}.}
\label{table:baseline_token_span_median>0}
\end{table}

\begin{table*}[h!]
    \centering
    \begin{tabular}{lcccccc}
        \toprule
        & $\mu$ & $\mu + \sigma$ & $\mu + 2\sigma$ & $\mu - \sigma$ & $\mu - 2\sigma$ & \textit{median} \\
        \midrule
        all & \cellcolor{lightgray}12.286 & 15.200 & 22.782 & 24.165 & 24.342 & 17.596 \\
        $>0$ & \cellcolor{lightgray}9.082 & 17.893 & 23.354 & 19.756 & 23.020 & \cellcolor{lightgray}10.265 \\
        
        \bottomrule
    \end{tabular}
    \caption{The averaged Euclidean distances between the methods' mean±stdev values for each threshold, and human preference (4±3). We analyse further the three thresholds visually indicated with a dark background that have nearest distance to human preference.}
    \label{tab:euclidean_distances}
\end{table*}

\begin{table*}[h!]
\centering
\begin{tabular}{l|ccc|ccc|ccc}
\multicolumn{1}{c}{} & \multicolumn{3}{c}{\textit{Stop words}} & \multicolumn{3}{c}{\textit{Punctuation}} & \multicolumn{3}{c}{\textit{POS}} \\
\toprule
\textbf{Comparison} & \textbf{$\chi^2$} & \textbf{$p$} & df & \textbf{$\chi^2$} & \textbf{$p$} & df & \textbf{$\chi^2$} & \textbf{$p$} & df \\ \toprule
PartSHAP vs LIME  & 0.0  & 1.0 & 1  & 0.0 & 1.0 & 1 & 0.0 & 1.0 & 4                               \\ \midrule
PartSHAP vs VanGrad & 1.247 & 0.264 & 1 & 0.255 & 0.614 & 1 & 2.580 & 0.630 & 4                               \\ \midrule
\cellcolor{lightgray}PartSHAP vs Grad×I & 7.642 & 0.006* & 1 & 3.763 & 0.052 & 1 & 10.361 & 0.035* & 4                               \\ \midrule
\cellcolor{lightgray}PartSHAP vs IntGrad & 6.155 & 0.013* & 1 & 2.216 & 0.137 & 1 & 9.578 & 0.048* & 4                               \\ \midrule
\cellcolor{lightgray}PartSHAP vs IntGrad×I & 3.611 & 0.057 & 1 & 2.962 & 0.085 & 1 & 5.219 & 0.266 & 4                               \\ \midrule
PartSHAP vs Human & 0.0  & 1.0 & 1 & 0.255 & 0.614 & 1 & 0.117 & 0.998 & 4                               \\ \midrule
LIME vs VanGrad & 0.886 & 0.347 & 1  & 0.820 & 0.365 & 1 & 2.580 & 0.630 & 4                               \\ \midrule
\cellcolor{lightgray}LIME vs Grad×I & 8.595 & 0.003* & 1 & 2.595 & 0.107 & 1 & 10.361 & 0.035* & 4                               \\ \midrule
\cellcolor{lightgray}LIME vs IntGrad  & 7.018 & 0.008* & 1 & 1.316 & 0.251 & 1 & 9.578 & 0.048* & 4                               \\ \midrule
\cellcolor{lightgray}LIME vs IntGrad×I & 4.287 & 0.038* & 1 & 1.920 & 0.166 & 1 & 5.219 & 0.266 & 4                               \\ \midrule
LIME vs Human & 0.0 & 1.0 & 1  & 0.820 & 0.365 & 1 & 0.117 & 0.998 & 4                               \\ \midrule
\cellcolor{lightgray}VanGrad vs Grad×I & 15.855 & <0.001* & 1 & 7.181 & 0.007* & 1 & 20.485 & <0.001* & 4                               \\ \midrule
\cellcolor{lightgray}VanGrad vs IntGrad & 13.747 & <0.001* & 1 & 5.158 & 0.023* & 1 & 19.476 & <0.001* & 4                               \\ \midrule
\cellcolor{lightgray}VanGrad vs IntGrad×I & 9.896 & 0.002* & 1 & 6.157 & 0.013* & 1 & 13.148 & 0.011* & 4                               \\ \midrule
VanGrad vs Human  & 0.886 & 0.347 & 1 & 0.0 & 1.0 & 1 & 2.635 & 0.621 & 4                               \\ \midrule
Grad×I vs IntGrad & 0.021 & 0.885 & 1 & 0.056 & 0.814 & 1 & 0.095 & 0.999 & 4                               \\ \midrule
Grad×I vs IntGrad×I & 0.536 & 0.464 & 1 & 0.0 & 1.0 & 1 & 1.544 & 0.819 & 4                               \\ \midrule
\cellcolor{lightgray}Grad×I vs Human & 8.595 & 0.003* & 1 & 7.181 & 0.007* & 1 & 10.212 & 0.037* & 4                               \\ \midrule
IntGrad vs IntGrad×I & 0.195 & 0.659 & 1 & 0.0 & 1.0 & 1 & 1.242 & 0.871 & 4                               \\ \midrule
\cellcolor{lightgray}IntGrad vs Human & 7.018 & 0.008* & 1 & 5.158 & 0.023* & 1 & 9.381 & 0.052 & 4                               \\ \midrule
\cellcolor{lightgray}IntGrad×I vs Human & 4.287 & 0.038* & 1 & 6.157 & 0.013* & 1 & 4.876 & 0.300 & 4                               \\ \bottomrule
\end{tabular}
\caption{Chi-Square test results for comparing different methods on their preference for stop words, punctuation and POS. Asterisk (*) indicates statistical significance at the 0.05 level. A dark background visually highlights the hypothesised Group 1 -- Group 2 comparisons.}
\label{tab:chi_squared_results}
\end{table*}

\end{document}